# Adaptive Initialization Method for K-means Algorithm

**Jie Yang**

jie.yang-4@student.uts.edu.au

Computational Intelligence and Brain Computer Interface Lab, Centre for AI, FEIT, University of Technology Sydney, Australia

**Yu-Kai Wang**

YuKai.Wang@uts.edu.au

Computational Intelligence and Brain Computer Interface Lab, Centre for AI, FEIT, University of Technology Sydney, Australia

**Xin Yao**

xiny@sustc.edu.cn

Shenzhen Key Laboratory of Computational Intelligence, Department of Computer Science and Engineering, Southern University of Science and Technology, Shenzhen 518055, China, and CERCIA, School of Computer Science, University of Birmingham, Birmingham B15 2TT, U.K.

**Chin-Teng Lin**

Chin-Teng.Lin@uts.edu.au

Computational Intelligence and Brain Computer Interface Lab, Centre for AI, FEIT, University of Technology Sydney, Australia

**Abstract**

The K-means algorithm is a widely used clustering algorithm that offers simplicity and efficiency. However, the traditional K-means algorithm uses the random method to determine the initial cluster centers, which make clustering results prone to local optima and then result in worse clustering performance. Many initialization methods have been proposed, but none of them can dynamically adapt to datasets with various characteristics. In our previous research, an initialization method for K-means based on hybrid distance was proposed, and this algorithm can adapt to datasets with different characteristics. However, it has the following drawbacks: (a) When calculating density, the threshold cannot be uniquely determined, resulting in unstable results. (b) Heavily depending on adjusting the parameter, the parameter must be adjusted five times to obtain better clustering results. (c) The time complexity of the algorithm is quadratic, which is difficult to apply to large datasets. In the current paper, we proposed an adaptive initialization method for the K-means algorithm (AIMK) to improve our previous work. AIMK can not only adapt to datasets with various characteristics but also obtain better clustering results within two interactions. In addition, we then leverage random sampling in AIMK, which is named as AIMK-RS, to reduce the time complexity. AIMK-RS is easily applied to large and high-dimensional datasets. We compared AIMK and AIMK-RS with 10 different algorithms on 16 normal and six extra-large datasets. The experimental results show that AIMK and AIMK-RS outperform the current initialization methods and several well-known clustering algorithms. Furthermore, AIMK-RS can significantly reduce the complexity of applying it to extra-large datasets with high dimensions. The time complexity of AIMK-RS is $O(n)$.

# 1. Introduction

The clustering algorithm is a classic algorithm in the field of data mining. It is used in virtually all natural and social sciences and has played a central role in various fields such as biology, astronomy, psychology, medicine, and chemistry (Shah and Koltun, 2017). For example, in the commercial field, Horng-Jinh Chang et al. proposed an anticipation model of potential customers' purchasing behavior based on clustering analysis (Chang et al., 2007). In the biology field, clustering is of central importance for the analysis of genetic data, as it is used to identify putative cell types (Kiselev et al., 2019). In addition, the applications of the clustering algorithm also include image segmentation, object or character recognition (Dorai and Jain, 1995), (Connell and Jain, 1998) and data reduction (Huang, 1997; Jiang et al., 2014). The clustering algorithm mainly includes hierarchy-based algorithms, partition-based algorithms, density-based algorithms, model-based algorithms and grid-based algorithms (Saxena et al., 2017).

The K-means algorithm is widely used because of its simplicity and efficiency (MacQueen, 1967). It is a classic partition-based clustering algorithm. However, the traditional K-means algorithm uses the random method to determine the initial cluster centers, which make clustering results prone to local optima and then result in worse clustering performance. To overcome this disadvantage, many improved methods have been proposed. However, providing an optimal partition is an N-P hard problem under a specific metric (Redmond and Heneghan, 2007).

Forgy randomly selected K points from the data as the initial cluster centers without a theoretical basis, and the final clustering results more easily fell into a local optimum (FORGY, 1965). Jancey's method assigned a randomly generated synthetic point from the data space to each initial clustering center (Jancey, 1966). However, some of these centers may be quite distant from any of the points, which might lead to the formation of empty clusters. Mac Queen proposed using the first K points in the dataset as the initial centers. The disadvantage of this approach is that the method is extremely sensitive to data order (MacQueen, 1967). In addition, the above methods do not take into account the characteristics of data distribution, using randomly generated points or synthetic points as the initial cluster centers, resulting in poor and unstable clustering results (Yang et al., 2017). Selecting clustering centers is actually selecting the representative points for specific classes. The density of data points can be used to measure the representativeness of points. Redmond et al. estimated the density distribution of the dataset by constructing a Kd-tree (Redmond and Heneghan, 2007), but its density calculation method was unreasonable (Wang et al., 2009). Geng Zhang et al. proposed an initialization method based on density Canopy with complexity $O(n^2)$ (Zhang et al., 2018). In addition, Cao et al. used the neighborhood-based rough set mode to measure the representativeness of the points to generate the initial cluster centers, but the method was sensitive to parameters (Cao et al., 2009). Khan et al. calculated the representative points from the dimensions of the data points based on the principle of data compression (Khan and Ahmad, 2004). The overall effect of this method is good, but its complexity is positively related to the dimensionality of the data and is not applicable to high-dimensional data. Based on the minimum spanning tree (MST), Jie et al. selected representative points, which are also called skeleton points, from the datasets and then regarded some skeleton points that are far away from each other as the final initial cluster centers (Yang et al., 2016). However, the complexity of this method is quadratic. In addition to the density of the data points, the distance between the data points is also regarded as one of the criteria for selecting the initial cluster centers. Gonzalez proposed a maximin method; the idea is to select the data points, which are as far as possible from each other, as the initial cluster centers, to make the cluster centers more evenly dispersed in each class (Gonzalez, 1985). However, this method has strong randomness, resulting in unstable clustering results. David Arthur et al. proposed K-means++ (Arthur and Vassilvitskii, 2007), which has disadvantages similar to the maximin method. For example, K-means++ will result in unstable clustering results because of the randomly selected first cluster center, or it may generate no representative initial cluster centers. To obtain better clustering results, some methods consider both the representativeness of data points and the distance between data points. A. Rodriguez et al. proposed a new clustering algorithm based on density peaks and proposed a method to generate cluster centers based on both density and distance (Rodriguez and Laio, 2014). Although this method is not leveraged to initialize the K-means algorithm, it has an extremely important enlightening significance. Numerous improved initialization methods have not yet been widely applied. Moreover, all the above mentioned methods cannot dynamically adapt to datasets with various characteristics (Yang et al., 2017).

Jie et al. proposed a K-means initialization method based on hybrid distance, which has been proven to dynamically adapt to datasets with various characteristics (Yang et al., 2017). The method considers both the density and the distance and uses a parameter to adjust the proportion of the two factors. They also proposed an internal clustering validation index, named the clustering validation index based on the neighbors (CVN), to select the optimal clustering results. However, this method also has shortcomings, such as (a) when calculating density, the threshold cannot be uniquely determined, resulting in unstable results. (b) Heavily depending on adjusting the parameter, the parameter must be adjusted five times to obtain better clustering results. (c) In some cases, the CVN index values calculated using different parameter settings are equal. At this time, CVN cannot be used to select better clustering results. (d) The time complexity of the algorithm is $O(n^2)$, which is difficult to apply to large datasets.

In this paper, we proposed an adaptive initialization method for the K-means algorithm (AIMK), which not only adapts to datasets with various characteristics but also requires only two runs to obtain better clustering results. In addition, we proposed the AIMK-RS based on random sampling to reduce the time complexity of the AIMK to $O(n)$. AIMK-RS is easily applied to large and high-dimensional datasets. First, we proposed a new threshold to calculate the density of the data points based on MST. In addition, after using the new threshold, we found that we only need to adjust the parameter twice, and we can obtain better clustering results. Finally, we applied random sampling to AIMK to obtain the AIMK-RS, whose time complexity is only $O(n)$.

This paper is organized as follows. In the Proposed Algorithm section, the new initialization method for K-means based on skeleton points is presented. In the Experiments and Results section, the simulation and the results are presented and discussed. Finally, in the Conclusion section, the relevant conclusions are drawn.

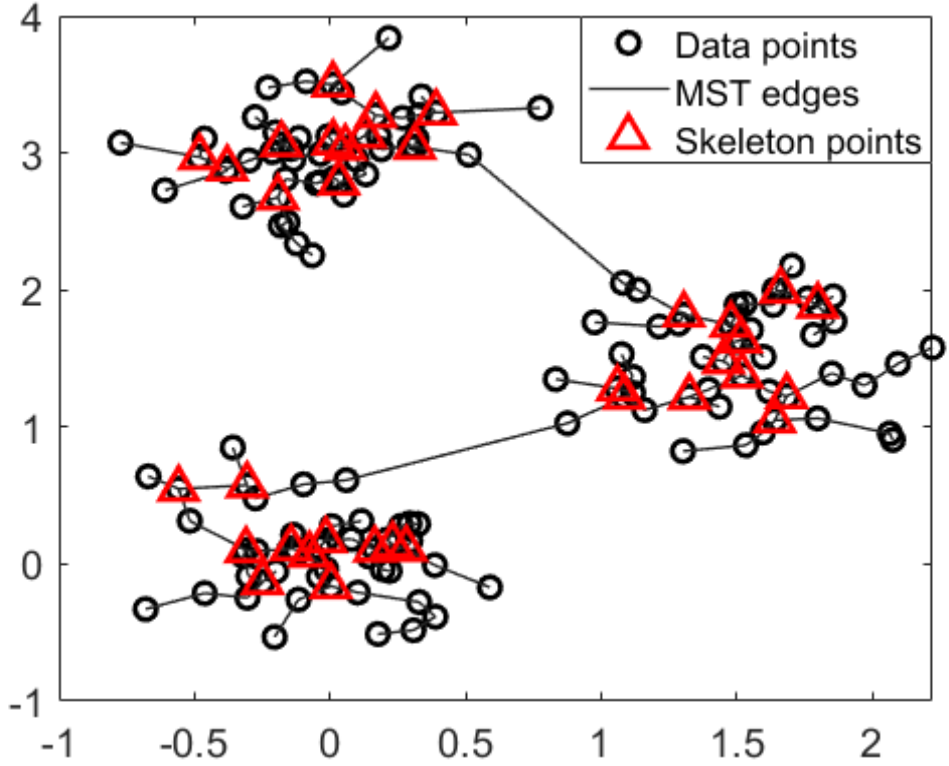

Fig. 1. We generated a synthetic dataset, then constructed the MST and calculated the skeleton points according to Definitions 2.1-2.3; they are enclosed by the triangles. As shown, the skeleton points are a type of compressed representation based on the characteristic of the dataset.

## 2. Adaptive initialization method

In this section, we described the algorithm for selecting the initial cluster centers in detail. First, several concepts involving this algorithm are presented.

### 2.1 Skeleton point

In a previous study, Jie et al. proposed a new compressed representation, named skeleton points, from the original datasets based on an MST and regarded them as candidates for cluster centers (Yang et al., 2016). In contrast, we leveraged the skeleton points to determine the threshold for calculating the density of data points because they can reflect the characteristics of the datasets to some extent. In the beginning, we introduce how to construct an MST using the original dataset.

Let X be a dataset with K clusters and n data points; that is, $X = \{x_i | x_i \in R^p, i = 1,2,\ldots,n\}$. To apply the MST to obtain the skeleton points, dataset X should be represented by the undirected complete weighted graph $G = (V, E)$, where $V = \{v_1, v_2, ..., v_n\}$, $|E| = \frac{n(n-1)}{2}$. Each data point $x_i$ in dataset X corresponds to a vertex $v_i \in V$ in graph G, and there is a one-to-one correspondence between data point $x_i$ (i = 1, 2..., n) and vertex $v_i$ (i = 1, 2..., n). The number of vertices in graph G is equal to the number of data points $x_i$ in dataset X. Edge weights between any two vertices are distance between the corresponding two data points.

The MST of G can be generated by using the Prim algorithm(Prim, 1957), which can be described as performing the following steps:

Step 1: Pick any vertex $v_i$ from graph G to be the root of the tree.

Step 2: Grow the tree by one edge: of the edges that connect the tree to vertices not yet in the tree, find the minimum-weight edge from G and transfer it to the tree.

Step 3: Repeat Step 2 (until the tree contains all vertices in graph G).

Through the above steps, we obtained an MST from the original datasets, and then we introduced how to obtain skeleton points from the MST. First, we introduce a concept, named the number of adjacent data points, and then we use this concept to obtain skeleton points.

Let $T = (V, E_T)$ be a minimum spanning tree of $G = (V, E)$, where $V = \{v_1, v_2, ..., v_n\}$, $E_T = \{e_1, e_2, ..., e_{n-1}\}, e_i \in E(G)$.

**Definition 2.1** (Number of adjacent data points) Let $U_i$ be the set of vertices of T with degree i and $W_i$ be the complementary set of $U_i$, that is, $W_i = V \backslash U_i$. For $U_i$, the number of adjacent data points, denoted as $f_i$, is the number of vertex in $W_i$ being adjacent to vertex in $U_i$.

Note that only add 1 to $f_i$ under the circumstance of one vertex in $W_i$ being adjacent to more than one vertex in $U_i$.

**Lemma 2.1** If anyone vertex in $W_i$ is adjacent to one and only one vertex in $U_1$, then $f_1 = |U_1|$, otherwise $f_1 < |U_1|$.

The proof of this lemma is obvious.

Now, we introduce how to leverage the number of adjacent data points $f_i$ to obtain the skeleton points.

**Definition 2.2** (Skeleton Point) Suppose the maximum degree of T be m; then, $V = U_1 \cup U_2 \cup ... \cup U_m$. Let $F = \arg\max_i f_i$. The skeleton points, denoted as S, are the vertices of T with the degree being greater than or equal to F. Therefore, $S = U_F \cup U_{F+1} \cup ... \cup U_m$.

We generated a synthetic dataset, then constructed the MST and calculated the skeleton points according to the Definition 2.1-2.3, which are enclosed by the triangles, as shown in Fig. 1. Next, we introduced the threshold for calculating the density of data points.

## 2.2 Threshold

**Definition 2.3** (Threshold) In $T = (V, E_T)$, suppose the number of skeleton points S is s; if the maximum weights of adjacent edges of each skeleton point can be denoted as $\{w_1, w_2, ..., w_s\}$, then we defined a threshold as

$$\text{Thr} = \frac{\sum_{i=1}^{s} w_i}{s} \tag{1}$$

In an MST, the adjacent edge weights of vertices can reflect the distribution characteristics of the area where the vertices are located. While vertices contain a large number of unimportant points or outliers, we only focus on the skeleton points. In summary, when calculating the threshold, we only consider the adjacent edge weights of the skeleton points, and the mean value of the maximum weights of adjacent edges of each skeleton point is taken as the threshold.

## 2.3 Density of vertices

In the following section, we introduce how to calculate the density of data points using the threshold *Thr*. We first constructed a Thr-Connected Graph.

**Definition 2.4** (Thr-Connected Graph) In dataset X, if $d(x_i, x_j) \leq Thr$, then add an edge between data points $x_i$ and $x_j$; this is called a Thr-Connected Graph (TCG), where $d(x_i, x_j)$ is the distance between data point $x_i$ and $x_j$. Each data point $x_i$ in dataset X corresponds to a vertex $v_i \in V$ in graph TCG.

**Definition 2.5** (density of $v_i$) In TCG, the mean distance between the vertex $v_i$ and the vertices adjacent to vertex $v_i$, denoted as $v_j$, is

$$D(v_i) = \frac{1}{k} \sum_{v_{i,j} \in V} d(v_i, v_j) \tag{2}$$

where k is the number of vertices $v_j$.

Suppose $v^k = \{v_i |$ the number of vertices adjacent to vertex $v_i$ is $k\}$, $D_{max}^k =$

$\max\limits_{v_i \in v^k} D(v_i)$, and $D^k_{min} = \min\limits_{v_i \in v^k} D(v_i)$; then, the density of $v_i$ is

$$\rho_i = \begin{cases} 0, & k = 0 \\ k + \dfrac{D^k_{max} - D(v_i)}{D^k_{max} - D^k_{min} + \varepsilon}, & k \neq 0 \end{cases} \quad (3)$$

To make $\dfrac{D^k_{max} - D(v_i)}{D^k_{max} - D^k_{min}} < 1$, we added infinite decimal $\varepsilon$ to its denominator, where $\varepsilon \to 0+$.

## 2.4 Hybrid distance

If the distance among the initial cluster centers is small, it is easy to make the K-means algorithm fall into a local optimum. However, if only the distance factor is considered, it is possible to use the outlier as the initial cluster center. Jie. et al. proposed a new distance, a hybrid distance, to solve this problem (Yang et al., 2017). Hybrid distance considers the distance and density of the cluster centers at the same time so that the selected cluster centers are far away from each other and have a higher density.

**Definition 2.6** (Hybrid distance between $v_i$) In TCG, suppose $d_{max} = \max\limits_{1 \leq i,j \leq n, i \neq j}\left(d(v_i, v_j)\right)$, $d_{min} = \min\limits_{1 \leq i,j \leq n, i \neq j}\left(d(v_i, v_j)\right)$, $P_{max} = \max\limits_{1 \leq i,j \leq n, i \neq j}(\rho_i + \rho_j)$, and $P_{min} = \min\limits_{1 \leq i,j \leq n, i \neq j}(\rho_i + \rho_j)$; the hybrid distance between the vertex $v_i$ and $v_j$ is

$$H(v_i, v_j) = \lambda \left[\dfrac{d(v_i,v_j) - d_{min}}{d_{max} - d_{min}}\right]^2 + (1-\lambda)\left[\dfrac{(\rho_i + \rho_j) - P_{min}}{P_{max} - P_{min}}\right]^2, \quad (4)$$

where λ is a hyperparameter, normally set by 0 or 1; this is explained in detail in the following section.

## 2.5 Algorithm for determining the initial cluster centers

Now we present the algorithm for determining the initial cluster centers based on the above-defined concepts. The details are as follows:

**Step 1:** Input: the dataset X and the number of clusters NC.

**Step 2:** Calculate distances $d(x_i, x_j)$ between all pairs of data points. Represent dataset X as the undirected complete weighted graph $G = (V, E)$ and construct the MST using the Prim algorithm, then calculate the threshold *Thr* by Definition 2.1-2.3.

**Step 3:** Construct the TCG by Definition 2.4, calculate the density of every vertex $\rho_i$ by Definition 2.5 and the sum of densities $\rho_i + \rho_j$ between all pairs of vertices.

**Step 4:** Calculate the hybrid distance $H(v_i, v_j)$ between all pairs of vertices according to Definition 2.6.

**Step 5:** Suppose a set of initial cluster centers $C = \Phi$. Find $v_i \in V$, which satisfies $\rho_i = \max\limits_{1 \leq k \leq n} \rho_k$. Then, the first initial cluster center $C_1 = v_i$, $C = \{C_1\}$, $V = V - v_i$.

**Step 6:** Find $v_i \in V$ which satisfies $i = \arg\max\limits_{v_k \in V} H(C_1, v_k)$. Then, the second initial cluster center $C_2 = v_i$, $C = \{C_1, C_2\}$, $V = V - v_i$.

**Step 7:** Find $v_i \in V$ which satisfies $i = \arg\max\limits_{v_k \in V}\left(\min\limits_{C_j \in C}\left(H(C_j, v_k)\right)\right)$. Then, $C = C \cup \{v_i\}$, $V = V - v_i$.

**Step 8:** Repeat Step 5 until $|C| = NC$.

**Step 9:** Output the set of initial cluster centers $C = \{C_1, C_2, \dots, C_{NC}\}$.

## 3 Experiments and Results

### 3.1 Datasets

We evaluated AIMK on 16 normal and six large and high-dimensional real-world datasets from the University of California at Irvine (UCI) (https://archive.ics.uci.edu/ml/datasets) and LIBSVM (https://www.csie.ntu.edu.tw/~cjlin/libsvmtools/datasets/), including Breast-cancer, Shuttle, Pendigits, Colon-cancer, Zoo, Haberman, Svmguide2, Wine, Ionosphere, Leukemia, Gisette, Splice, Svmguide4, Liver-disorders, Soybean-small, Balance-scale, Ijcnn1, Phishing, Protein, Mushrooms, SensIT Vehicle (seismic), SensIT Vehicle (combined). The description of the datasets is as shown in Table 1.

Table 1. Description of the 22 datasets

| Dataset | Number of Instances | Number of Attributes | Number of Classes |
|---|---|---|---|
| Breast-cancer | 683 | 10 | 2 |
| Shuttle | 14,500 | 9 | 7 |
| Pendigits | 3,498 | 16 | 10 |
| Colon-cancer | 62 | 2000 | 2 |
| Zoo | 101 | 16 | 7 |
| Haberman | 306 | 3 | 2 |
| Svmguide2 | 391 | 20 | 3 |
| Wine | 178 | 13 | 3 |
| ionosphere | 351 | 34 | 2 |
| Leukemia | 34 | 7129 | 2 |
| Gisette | 1,000 | 5,000 | 2 |

| Splice | 2,175 | 60 | 2 |
|---|---|---|---|
| Svmguide4 | 312 | 10 | 6 |
| Liver-disorders | 200 | 5 | 2 |
| Soybean-small | 47 | 35 | 4 |
| Balance-scale | 625 | 4 | 3 |
| Ijcnn1 | 91,701 | 22 | 2 |
| Phishing | 11,055 | 68 | 2 |
| Protein | 6,621 | 357 | 3 |
| Mushrooms | 8124 | 112 | 2 |
| SensIT Vehicle (seismic) | 19,705 | 50 | 3 |
| SensIT Vehicle (combined) | 19,705 | 100 | 3 |

**3.2 Baselines**

We compared AIMK to 10 baselines. For the sake of fairness, these baselines not only include initialization methods for K-means, such as K-means (MacQueen, 1967), K-means++ (Arthur and Vassilvitskii, 2007), the method initializing K-means using kd-trees (KT) (Redmond and Heneghan, 2007), the MST-based method for initializing K-means (MSTI) (Yang et al., 2016), and the initialization method based on hybrid distance for K-means (HD) (Yang et al., 2017), but also include some widely used clustering algorithms, such as K-medoids (Kaufman and Rousseeuw, 2009), clustering by fast search and find of density peaks (SFDP) (Rodriguez and Laio, 2014), fuzzy C-means clustering (FCM) (Bezdek et al., 1984), single-linkage hierarchical clustering (SH) (Johnson, 1967), and self-tuning spectral clustering (SS) (Zelnik-manor and Perona, 2005). In addition, since the results of K-means, K-means++, K-medoids, FCM, and SS are not unique, we take the average performance of 10 runs as the real baseline performance. SFDP has a hyperparameter dc, ranging from 1%~2% (Rodriguez and Laio, 2014). We take the average performance while the hyperparameter equals 1%, 1.1%, 1.2%, 1.3%, 1.4%, 1.5%, 1.6%, 1.7%, 1.8%, 1.9%, and 2% because of the sensitivity of the parameter. Note that all the baselines and AIMK used Euclidean distance as a metric, even if the attribute values of the two datasets, Soybean-small and Balance-scale, are categorical.

**3.3 Validation Indices**

To evaluate the performance of clustering algorithms, we exploit three widely used external clustering validation indices: Accuracy (ACC), Rand Index (RI), and F-measure. These indices are defined as follows:

$$\text{ACC} = \frac{\sum_{i=1}^{NC} P_i}{n} \tag{5}$$

$$\text{RI} = \frac{TP+TN}{TP+FP+FN+TN} \tag{6}$$

$$\text{Precision} = \frac{TP}{TP+FP} \tag{7}$$

$$\text{Recall} = \frac{TP}{TP+FN} \tag{8}$$

$$\text{F}-\text{measure} = \frac{2*Precision*Recall}{Precision+Recall} \tag{9}$$

where *n* denotes the number of objects. *NC* is the number of clusters. $P_i$ is the number of objects that are correctly assigned. TP means true positive, FP means false positive, FN means false negative, and TN means true negative (Powers, 2011).

### 3.4 Sensitivity of λ

To analyze the sensitivity of the parameter λ, we ran AIMK on 16 datasets: Breast-cancer, Shuttle, Pendigits, Colon-cancer, Zoo, Haberman, Svmguide2, Wine, Ionosphere, Leukemia, Gisette, Splice, Svmguide4, Liver-disorders, Soybean-small, and Balance-scale while λ is set as 0, 0.25, 0.5, 0.75, 1. Then, we used ACC, RI, and F-measure to evaluate the performance of AIMK on each dataset. The results are listed in Tables 2–4. The optimal results for the corresponding index are denoted in bold. As the results show, when λ is set as 0 or 1, the optimal results in each validation index can be always obtained in each dataset. The HD algorithm is required to run five times to obtain a better clustering result (Yang et al., 2017), but AIMK can obtain a better result with only two runs. Therefore, in subsequent experiments, we only consider the results of AIMK when λ equals 0 or 1.

Table 2. AIMK run on 16 datasets, measured by ACC.

| λ \ Dataset | λ=0 | λ=0.25 | λ=0.5 | λ=0.75 | λ=1 |
|---|---|---|---|---|---|
| Breast-cancer | .6032 | .6032 | .6471 | .6471 | **.6471** |
| Shuttle | .4598 | .4598 | .4598 | .5994 | **.8327** |
| Pendigits | **.7424** | .6755 | .6575 | .6161 | .5780 |
| Colon-cancer | **.8710** | .8710 | .8710 | .5484 | .6129 |
| Zoo | .6436 | .6436 | .6436 | .7624 | **.8416** |
| Haberman | .5000 | .5000 | .5000 | .7582 | **.7582** |
| Svmguide2 | .4501 | .4501 | .4501 | .5985 | **.5985** |
| Wine | **.7022** | .7022 | .7022 | .7022 | .5730 |

| Ionosphere | **.7123** | .7123 | .7123 | .6439 | .6439 |
| Leukemia | .5882 | .5882 | .5882 | .6176 | **.6176** |
| Gisette | .6650 | .6650 | .6700 | .6700 | **.6700** |
| Splice | .5160 | .6560 | .6560 | .6560 | **.6560** |
| Svmguide4 | **.2967** | .2967 | .2600 | .2933 | .2633 |
| Liver-disorders | **.7448** | .7448 | .7448 | .7103 | .7103 |
| Soybean-small | **1** | 1 | 1 | .7234 | .7447 |
| Balance-scale | .5488 | .6016 | .6016 | .5264 | **.6144** |

Table 3. AIMK run on 16 datasets, measured by RI.

| λ \ Dataset | λ=0 | λ=0.25 | λ=0.5 | λ=0.75 | λ=1 |
|---|---|---|---|---|---|
| Breast-cancer | .5206 | .5206 | .5426 | .5426 | **.5426** |
| Shuttle | .5600 | .5600 | .5600 | .5799 | **.7578** |
| Pendigits | **.9214** | .9102 | .9074 | .9001 | .8852 |
| Colon-cancer | **.7715** | .7715 | .7715 | .4966 | .5177 |
| Zoo | .7580 | .7580 | .7580 | .9115 | **.9228** |
| Haberman | .4984 | .4984 | .4984 | .6321 | **.6321** |
| Svmguide2 | **.5669** | .5669 | .5669 | .5622 | .5622 |
| Wine | **.7187** | .7187 | .7187 | .7187 | .6919 |
| Ionosphere | **.5889** | .5889 | .5889 | .5401 | .5401 |
| Leukemia | .5009 | .5009 | .5009 | .5134 | **.5134** |
| Gisette | .5540 | .5540 | .5574 | .5574 | **.5574** |
| Splice | .5000 | .5482 | .5482 | .5482 | **.5482** |
| Svmguide4 | **.7219** | .7219 | .6999 | .7205 | .6698 |
| Liver-disorders | **.6172** | .6172 | .6172 | .5856 | .5856 |
| Soybean-small | **1** | 1 | 1 | .8316 | .8335 |
| Balance-scale | .5741 | .6724 | .6724 | .5959 | **.6866** |

Table 4. AIMK run on 16 datasets, measured by F-measure.

| λ \ Dataset | λ=0 | λ=0.25 | λ=0.5 | λ=0.75 | λ=1 |
|---|---|---|---|---|---|
| Breast-cancer | .5852 | .5852 | .7027 | .7027 | **.7027** |
| Shuttle | .5016 | .5016 | .5016 | .5600 | **.8430** |
| Pendigits | **.6180** | .5966 | .5902 | .5705 | .5317 |
| Colon-cancer | **.7803** | .7803 | .7803 | .5118 | .6584 |
| Zoo | .5999 | .5999 | .5999 | .8051 | **.8297** |
| Haberman | .5482 | .5482 | .5482 | .7290 | **.7290** |
| Svmguide2 | .4283 | .4283 | .4283 | .5255 | **.5255** |
| Wine | .5835 | .5835 | .5835 | .5835 | **.5956** |
| Ionosphere | .6049 | .6049 | .6049 | .6999 | **.6999** |
| Leukemia | .5156 | .5156 | .5156 | .6625 | **.6625** |
| Gisette | .5788 | .5788 | .6062 | .6062 | **.6062** |
| Splice | **.6662** | .5551 | .5551 | .5545 | .5545 |
| Svmguide4 | .2029 | .2029 | .2060 | .2000 | **.2178** |
| Liver-disorders | **.6798** | .6798 | .6798 | .6754 | .6754 |
| Soybean-small | 1 | 1 | 1 | .6566 | .6617 |
| Balance-scale | .4601 | .5721 | .5721 | .4719 | **.5901** |

## 3.5 Impact of threshold *Thr*

To explain more clearly how to use skeleton points to determine the threshold *Thr*, we perform experiments on four representative datasets: Pendigits, Shuttle, Wine, and Gisette, whose data size and dimensions are from small to large and low to high, respectively. We run AIMK when *Thr* is set as the mean value of the maximum weights, mean weights, and minimum weights of adjacent edges of each skeleton point. Meanwhile, λ is set as 0 or 1, and the final results are shown as both sides of the slash "/", respectively. We used ACC to evaluate the results of each run. The results are shown in Table 5. For each dataset, the optimal results can be obtained only when *Thr* is set as the maximum weights of adjacent edges of each skeleton point. Therefore, it is more reasonable to set *Thr* as the maximum weights of adjacent edges of each skeleton point.

Table 5. The impact of threshold *Thr* on clustering performance.

| Dataset \ *Thr* | Min | Mean | Max |
|---|---|---|---|
| Pendigits | .5780/ .5780 | .5780/.5780 | **.7424**/.5780 |
| Shuttle | .7530/.7530 | .7530/.7530 | .4598/**.8327** |
| Wine | .5730/.5730 | .5730/.5730 | **.7022**/.5730 |
| Gisette | .5010/.5010 | .5010/.5010 | .6650/**.6700** |

## 3.6 Comparison with baselines

We compared AIMK (λ is set as 0 or 1) with 10 baselines on 16 normal real-world datasets. ACC, RI, and F-measure are exploited to evaluate the performance of each baseline on each dataset. The results are listed in Tables 6–8. The optimal results for the corresponding dataset are denoted in bold. We use the average rank to measure the final performance of each baseline across datasets. The rank means the rank number of each row sorted in descending order. If there are the same results from two different algorithms, their ranks are equal.

According to Tables 6-8, AIMK (set λ as 0 or 1) achieves the best performance on 13, 11, and 8 of the 14 datasets when measured by ACC, RI, and F-measure, respectively. Moreover, it can be seen from the ranks that AIMK is obviously superior to the other 10 baselines, no matter which validation index we use.

Furthermore, according to Table 6, AIMK achieves the highest ACC rank compared with the other 10 baselines. The rank of AIMK 1.125 is much higher than the rank of HD 4.188, which achieves the second-highest ACC rank. FCM achieves the lowest ACC rank, at just 7.438. HD is the best-performing initialization method for K-means in addition to AIMK in Table 7, whose rank is 4.188. According to Table 7, AIMK achieves the highest RI rank compared with the other 10 baselines. The rank of AIMK 1.312 is much higher than the rank of HD 4.312, which achieves the second-highest RI rank. SH achieves the lowest RI, at just 7.562. HD is still the best-performing initialization method for K-means in addition to AIMK in Table 8, whose rank is 4.312. According to Table 8, AIMK still achieves the highest F-measure rank compared with the other 10 baselines. The rank of AIMK 1.500 is higher than the rank of SH 2.875, which achieves the second-highest F-measure rank. FCM achieves the lowest F-measure, which is just 9.062. MSTI is the best-performing initialization method for K-means in addition to AIMK in Table 9, whose rank is 5.125.

Table 6. Results of all algorithms on 16 real-world datasets, measured by ACC.

| Algorithm / Dataset | K-means | K-means++ | KT | MSTI | HD | K-medoids | SFDP | FCM | SH | SS | AIMK (λ=0) | AIMK (λ=1) |
|---|---|---|---|---|---|---|---|---|---|---|---|---|
| Breast-cancer | .6032 | .6252 | **.6471** | .6032 | **.6471** | **.6471** | .5928 | .6032 | **.6471** | **.6471** | .6032 | **.6471** |
| Shuttle | .4384 | .4588 | .6590 | .5858 | **.8327** | .4683 | .4130 | .4002 | .7914 | .3386 | .4598 | **.8327** |
| Pendigits | .6479 | .6609 | .5895 | .6795 | .5780 | .6553 | .6832 | .6048 | .1123 | .6685 | **.7424** | .5780 |
| Colon-cancer | .5613 | .6194 | .7742 | .5161 | .6129 | .6258 | .6818 | .5758 | .6290 | .5226 | **.8710** | .6129 |
| Zoo | .6644 | .7188 | .7327 | .7921 | **.8416** | .7921 | .5644 | .5752 | .6238 | .5406 | .6436 | **.8416** |
| Haberman | .5408 | .5121 | .5000 | .5196 | .5196 | .5196 | .5698 | .5098 | .7386 | .5196 | .5000 | **.7582** |
| Svmguide2 | .4624 | .4737 | .4680 | .5115 | .4655 | .4680 | .4076 | .5151 | .5703 | .4760 | .4501 | **.5985** |
| Wine | .6893 | .6640 | .5730 | .7022 | .7022 | .6820 | **.7079** | .6854 | .3764 | **.7079** | .7022 | .5730 |
| Ionosphere | .7103 | .7100 | .7094 | **.7123** | .7094 | .7094 | .5335 | .7094 | .6439 | **.7123** | **.7123** | .6439 |
| Leukemia | .5765 | .5882 | .5588 | .5294 | .5882 | .5294 | .5294 | .5294 | **.6176** | .5588 | .5882 | **.6176** |
| Gisette | .6538 | .6548 | .6540 | .6690 | .6650 | .6281 | .6300 | .6595 | .5010 | .6664 | .6650 | **.6700** |
| Splice | .6409 | .6539 | .6550 | .6540 | .6550 | .5990 | .5070 | .6283 | .5160 | .6476 | .5160 | **.6560** |
| Svmguide4 | .2720 | .2597 | .2633 | .2633 | .2867 | .2620 | .2500 | .2590 | .1967 | .2653 | **.2967** | .2633 |
| Liver-disorders | .7283 | .7269 | .7103 | .7103 | .7103 | .7034 | .6038 | .7241 | .6276 | .6745 | **.7448** | .7103 |

15| | | | | | | | | | | | | |
|---|---|---|---|---|---|---|---|---|---|---|---|---|
| Soybean-small | .7191 | .7319 | .7447 | .7660 | .7447 | .8085 | .8936 | .7234 | **1** | .7787 | **1** | .7447 |
| Balance-scale | .5144 | .5179 | .4400 | .5408 | **.6144** | .5363 | .5439 | .5245 | .4640 | .4104 | .5488 | **.6144** |
| **Rank** | 6.688 | 6.312 | 6.062 | 4.812 | 4.188 | 6.125 | 7.250 | 7.438 | 6.562 | 5.812 | **1.125** | |

Table 7. Results of all algorithms on 16 real-world datasets, measured by RI.

| Algorithm / Dataset | K-means | K-means++ | KT | MSTI | HD | K-medoids | SFDP | FCM | SH | SS | AIMK (λ=0) | AIMK (λ=1) |
|---|---|---|---|---|---|---|---|---|---|---|---|---|
| Breast-cancer | .5228 | .5338 | **.5426** | .5206 | **.5426** | **.5426** | .5179 | .5206 | **.5426** | **.5426** | .5206 | **.5426** |
| Shuttle | .5201 | .5567 | .5846 | .5802 | **.7578** | .5652 | .4847 | .5115 | .6520 | .4735 | .5600 | **.7578** |
| Pendigits | .9021 | .9148 | .8963 | .9079 | .8852 | .9098 | .9193 | .8869 | .1147 | .9165 | **.9214** | .8852 |
| Colon-cancer | .5101 | .5334 | .6446 | .4923 | .5177 | .5454 | .5618 | .5015 | .5256 | .4961 | **.7715** | .5177 |
| Zoo | .8283 | .8786 | .8618 | .8994 | **.9228** | .8953 | .7657 | .8386 | .7186 | .8088 | .7580 | **.9228** |
| Haberman | .4989 | .5122 | .4984 | .4991 | .4991 | .4991 | .5081 | .4986 | .6126 | .4991 | .4984 | **.6321** |
| Svmguide2 | .5621 | .5646 | .5544 | .5738 | .5532 | **.5812** | .4905 | .5585 | .4317 | .5610 | .5669 | .5622 |
| Wine | .7079 | .7049 | .6919 | .7187 | .7187 | .7172 | .7191 | .7105 | .3479 | **.7204** | .7187 | .6919 |
| Ionosphere | .5880 | .5870 | .5865 | **.5889** | .5865 | .5865 | .5054 | .5865 | .5401 | **.5889** | **.5889** | .5401 |
| Leukemia | .4955 | .4898 | .4920 | .4866 | .5009 | .4866 | .4866 | .4866 | **.5134** | .4920 | .5009 | **.5134** |
| Gisette | .5481 | .5534 | .5470 | .5567 | .5540 | .5298 | .5333 | **.5610** | .4995 | .5549 | .5540 | .5574 |
| Splice | .5467 | .5471 | .5476 | .5470 | .5476 | .5191 | .4996 | .5262 | .5000 | .5434 | .5000 | **.5482** |
| Svmguide4 | .7078 | .7010 | .6906 | .6702 | .7181 | .6895 | .7178 | .7208 | .1885 | .7159 | **.7219** | .6698 |
| Liver-disorders | .6064 | .5932 | .5856 | .5856 | .5856 | .5799 | .5186 | .5977 | .5293 | .5560 | **.6172** | .5856 |
| Soybean-small | .8286 | .8313 | .8335 | .8372 | .8335 | .8501 | .8982 | .8316 | **1** | .8417 | **1** | .8335 |
| Balance-scale | .5852 | .5888 | .5428 | .6171 | **.6866** | .5889 | .5801 | .6008 | .4329 | .5354 | .5741 | **.6866** |
| **Rank** | 6.750 | 5.812 | 6.250 | 5.312 | 4.312 | 5.688 | 7.312 | 6.750 | 7.562 | 5.750 | **1.312** | |

Table 8. Results of all algorithms on 16 real-world datasets, measured by F-measure.

| Algorithm / Dataset | K-means | K-means++ | KT | MSTI | HD | K-medoids | SFDP | FCM | SH | SS | AIMK (λ=0) | AIMK (λ=1) |
|---|---|---|---|---|---|---|---|---|---|---|---|---|
| Breast-cancer | .5969 | .6557 | **.7027** | .5852 | **.7027** | **.7027** | .5922 | .5852 | **.7027** | **.7027** | .5852 | **.7027** |
| Shuttle | .4291 | .4966 | .5909 | .5603 | **.8430** | .5149 | .4199 | .4134 | .7892 | .3535 | .5016 | **.8430** |
| Pendigits | .5679 | .5997 | .5499 | .5933 | .5316 | .5809 | **.6267** | .5264 | .1816 | .5912 | .6180 | .5317 |
| Colon-cancer | .5339 | .5625 | .6582 | .5176 | .6584 | .5587 | .6610 | .5151 | .6843 | .5096 | **.7803** | .6584 |
| Zoo | .6227 | .7171 | .6984 | .7608 | **.8297** | .7536 | .4601 | .5845 | .6169 | .5435 | .5999 | **.8297** |
| Haberman | .5492 | .5669 | .5482 | .5504 | .5480 | .5479 | .5945 | .5479 | **.7583** | .5480 | .5482 | .7290 |
| Svmguide2 | .4238 | .4274 | .4143 | .4436 | .4125 | .4487 | .4289 | .4543 | **.5986** | .4215 | .4283 | .5255 |
| Wine | .5883 | .5885 | **.5956** | .5835 | .5835 | .5858 | .5834 | .5728 | .4959 | .5859 | .5835 | **.5956** |
| Ionosphere | .6041 | .6032 | .6028 | .6049 | .6028 | .6024 | .5929 | .6028 | **.6999** | .6041 | .6049 | **.6999** |
| Leu | .5240 | .5530 | .4991 | .5727 | .5156 | .5017 | .4875 | .4875 | **.6625** | .4991 | .5156 | **.6625** |
| Gisette | .5772 | .5895 | .5860 | .6053 | .5788 | .6041 | .6157 | .5640 | **.6658** | .5583 | .5788 | .6062 |

16| | | | | | | | | | | | | |
|---|---|---|---|---|---|---|---|---|---|---|---|---|
| Splice | .5533 | .5538 | .5540 | .5539 | .5543 | .5215 | .5750 | .5272 | **.6662** | .5478 | **.6662** | .5545 |
| Svmguide4 | .1990 | .2012 | .2087 | .2171 | .1975 | .2064 | .1829 | .1908 | **.2835** | .1979 | .2029 | .2178 |
| Liver-disorders | .6753 | .6739 | .6654 | .6754 | .6754 | .6683 | .5815 | .6567 | **.6887** | .6001 | .6798 | .6754 |
| Soybean-small | .6745 | .6882 | .6617 | .6716 | .6617 | .6955 | .7925 | .6566 | **1** | .7080 | **1** | .6617 |
| Balance-scale | .4578 | .4629 | .4028 | .4991 | .5901 | .4724 | .4983 | .4827 | **.6016** | .3919 | .4601 | .5901 |
| **Rank** | 6.875 | 5.688 | 6.250 | 5.125 | 5.625 | 6.312 | 6.688 | 9.062 | 2.875 | 7.812 | **1.500** | |

## 3.7 Reducing complexity by sampling

Due to the time complexity $O(n^2)$, it is difficult to apply AIMK to large and high-dimensional datasets. To solve this problem, we consider random sampling to extract $\sqrt{n}$ samples from the original datasets, where $n$ means the number of samples of the dataset, and then use these samples as the input for AIMK. It is worth mentioning that to make the $\sqrt{n}$ samples fully express the characteristics of the original datasets, we recommend using random sampling to reduce complexity only when the number of clusters $K \ll n$. In this way, the time complexity of AIMK will be reduced to $O(n)$. AIMK after random sampling, denoted as AIMK-RS, is compared with two widely used initialization methods, K-means and K-means++, whose time complexity is also $O(n)$, on six large and high-dimensional datasets. ACC, RI, and F-measure are also exploited to evaluate the results. In addition, we take the average performance of 100 runs as the real performance of the AIMK-RS because it provides for more even sampling and can fully express the characteristics of the original datasets. The optimal results for the corresponding datasets are denoted in bold. The results are listed in Tables 9–11, and we can conclude that after random sampling, compared with two baselines, AIMK still achieves better performance.

Table 9. Large and high-dimensional datasets, measured by ACC.

| Algorithm / Dataset | K-means | K-means++ | AIMK-RS ($\lambda=0$) | AIMK-RS ($\lambda=1$) |
|---|---|---|---|---|
| Ijcnn1 | .7332 | .7472 | .6712 | **.8240** |
| Phishing | .5696 | .5715 | **.6208** | .5260 |
| Protein | .4252 | .4258 | .4173 | **.4576** |
| Mushrooms | .7918 | .8083 | .8027 | **.8241** |
| SensIT Vehicle (seismic) | .4546 | .4657 | .4425 | **.4855** |
| SensIT Vehicle (combined) | .5576 | .5598 | **.5636** | .5384 |

Table 10. Large and high-dimensional datasets, measured by RI.

| Algorithm / Dataset | K-means | K-means++ | AIMK-RS ($\lambda=0$) | AIMK-RS ($\lambda=1$) |
|---|---|---|---|---|
| Ijcnn1 | .6207 | .6367 | .5603 | **.7099** |

| Dataset | | | | |
|---|---|---|---|---|
| Phishing | .5309 | .5315 | **.5682** | .5020 |
| Protein | .4390 | .4307 | **.4476** | .3666 |
| Mushrooms | .6989 | .7206 | .7126 | **.7300** |
| SensIT Vehicle (seismic) | .5650 | .5653 | **.5658** | .5517 |
| SensIT Vehicle (combined) | .5941 | .5965 | **.5975** | .5709 |

Table 11. Large and high-dimensional datasets, measured by F-measure.

| Algorithm \ Dataset | K-means | K-means++ | AIMK-RS ($\lambda=0$) | AIMK-RS ($\lambda=1$) |
|---|---|---|---|---|
| Ijcnn1 | .7439 | .7581 | .6896 | **.8240** |
| Phishing | .5933 | .5937 | **.6046** | .5733 |
| Protein | .4780 | .4825 | .4683 | **.5268** |
| Mushrooms | .7253 | .7428 | .7364 | **.7595** |
| SensIT Vehicle (seismic) | .4041 | .4118 | .3949 | **.4374** |
| SensIT Vehicle (combined) | .4506 | .4513 | .4495 | **.4627** |

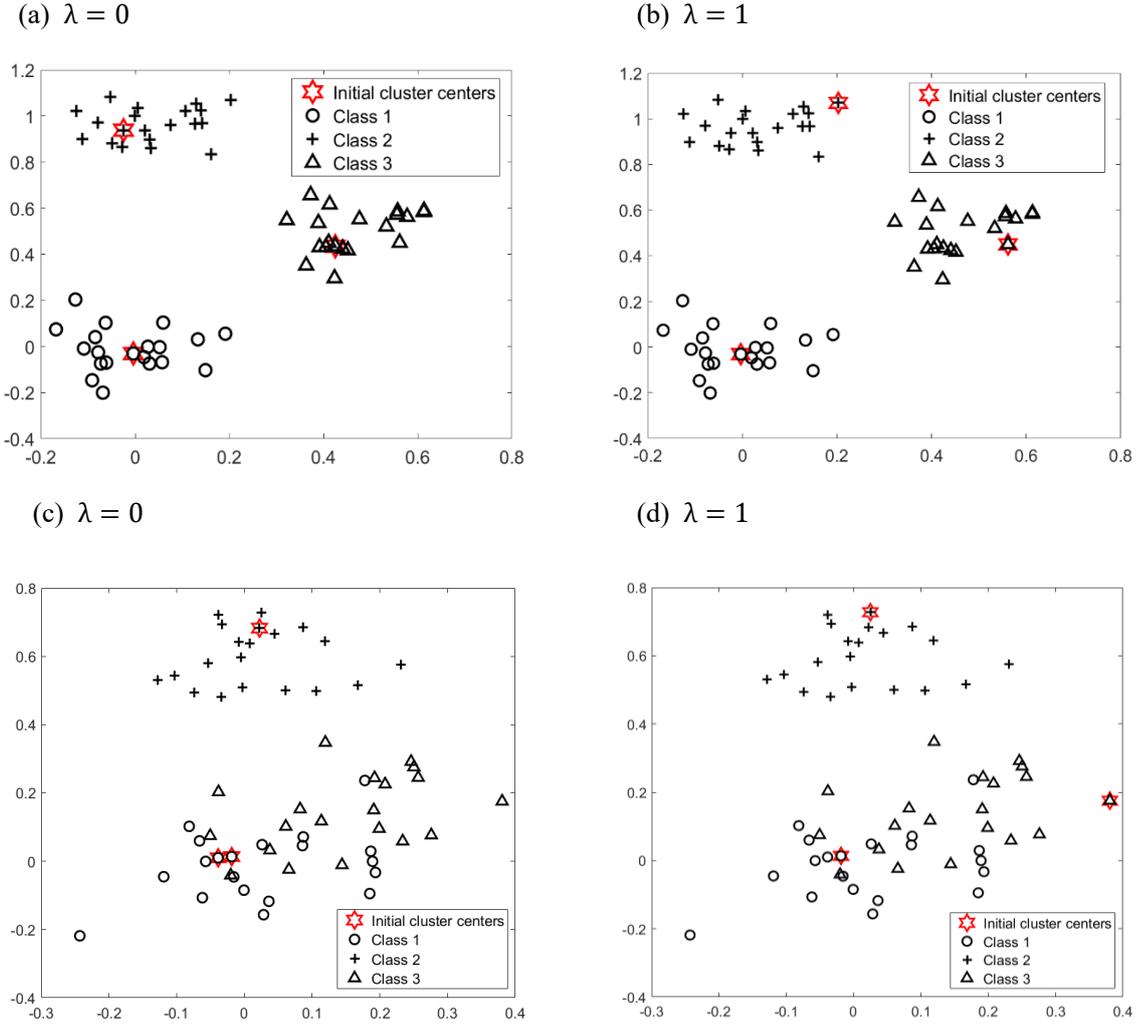

Fig. 2. To further illustrate the impact of parameter λ, we generated two types of datasets, (a)-(b) and (c)-(d), with different distribution from a mixture of three bivariate Gaussian densities. Class 1, Class 2, Class 3, and initial cluster centers are represented by different shapes: circle, cross, triangle, and star, respectively.

## 3.8 Choice of λ

After the above experiments, we can see that the parameter λ is crucial for the final clustering results. To further illustrate the impact of parameter λ, we generated two types of datasets with different distributions from a mixture of three bivariate Gaussian densities. Fig. 2(a)-Fig. 2(b) is given by

$$\tfrac{1}{3} Gaussian \begin{pmatrix}0\\0\end{pmatrix}\begin{pmatrix}0.01 & 0\\ 0 & 0.01\end{pmatrix} + \tfrac{1}{3} Gaussian \begin{pmatrix}0\\1\end{pmatrix}\begin{pmatrix}0.01 & 0\\ 0 & 0.01\end{pmatrix} + \tfrac{1}{3} Gaussian \begin{pmatrix}0.5\\0.5\end{pmatrix}\begin{pmatrix}0.01 & 0\\ 0 & 0.01\end{pmatrix},$$

Fig. 2(c)-Fig. 2(d) is given by

$$\frac{1}{3} Gaussian \begin{pmatrix} 0 \\ 0 \end{pmatrix} \begin{pmatrix} 0.01 & 0 \\ 0 & 0.01 \end{pmatrix} + \frac{1}{3} Gaussian \begin{pmatrix} 0 \\ 0.6 \end{pmatrix} \begin{pmatrix} 0.01 & 0 \\ 0 & 0.01 \end{pmatrix} + \frac{1}{3} Gaussian \begin{pmatrix} 0.15 \\ 0.15 \end{pmatrix} \begin{pmatrix} 0.01 & 0 \\ 0 & 0.01 \end{pmatrix}.$$

where Gaussian$(X, Y)$ is a Gaussian normal distribution with the mean X and the covariance matrix Y. We stimulate three clusters, namely, class 1, class 2 and class 3, which are represented by different shapes: circle (20 points), cross (20 points), and triangle (20 points), respectively. As shown in Fig. 2(a)-Fig. 2(d), we used AIMK to determine the initial cluster centers marked with star when λ is set as 0 and 1. In Fig. 2(a)-Fig. 2(b), when λ is equal to 0, the three cluster centers happen to be the centroid of three classes. When λ is equal to 1, only one cluster center is the centroid of class 1, and the other two cluster centers are just outliers in class 2 and class 3, respectively. In Fig. 2(c)-Fig. 2(d), when λ is equal to 0, two cluster centers are dropped in class 1, one cluster center is dropped in class 2, and no cluster center is dropped in class 3. However, when λ is equal to 1, three cluster centers happen to be dropped in three classes, and two of the three are outliers.

According to formula (4), when λ is equal to 0, only the top K points with higher density are selected as initial cluster centers. At this time, if all or most of these K initial cluster centers fall in K different classes, as shown in Fig. 2(a), then the initialization effect is better. However, for some datasets, such as overlapping datasets, shown as Fig. 2(c), the top K points with higher density cannot be distributed relatively evenly among K classes. Therefore, at this time, we need to consider the distance factor. According to formula (4), when λ is equal to 1, we only select the K points that are far apart from each other as initial cluster centers. At this time, all or most of these K initial cluster centers are more likely to be relatively evenly distributed among the K classes, as shown in Fig. 2(d).

## 3.9 Algorithm analysis

According to the 2.8 Algorithm for determining the initial cluster centers, the time complexity of AIMK was analyzed as follows. In Step 2, the time complexity of computation of the distance between all pairs of vertices and the Prim algorithm is $O(n^2)$, and the time complexity of calculation of the threshold *Thr* is $O(n)$. Construction of the TCG and calculation of the density of every vertex $\rho_i$ requires $O(n)$ in Step 3, and the computation of the sum of densities $\rho_i + \rho_j$ between all pairs of vertices requires $O(n)$. In Step 4, because the distance and the sum of densities between all pairs of vertices have been calculated in Step 2 and Step 3, the time complexity of the calculation of the hybrid distance $H(v_i, v_j)$ between all pairs of vertices is $O(n)$. Determination of the first and second initial cluster centers requires $O(n)$ in Step 5 and Step 6. In Steps 7 and 8, the remaining initial cluster centers are selected, in which the time complexity is less than $NC*n$ and approximately equal to $O(NC*n)$. Because normally the number of clusters $NC \ll n$, the entire time complexity of AIMK is $O(n^2)$. However, according to the 3.7 Reducing complexity by sampling part, the time complexity of AIMK can be reduced to $O(n)$ after random sampling, denoted as AIMK-RS. The time complexities of all baselines, AIMK and AIMK-RS are listed in Table 12.

Table 12. Comparison of time complexity of different clustering algorithms.

| Algorithm | Time Complexity |
| --- | --- |
| K-means | $O(n)$ |
| K-means++ | $O(n)$ |
| KT | $O(nlogn)$ |
| MSTI | $O(n^2)$ |
| HD | $O(n^2)$ |
| K-medoids | $O(n^2)$ |
| SFDP | $O(n^2)$ |
| FCM | $O(n)$ |
| Sing-linkage | $O(n^2 logn)$ |
| Self-tuning Spectral | $O(n^2)$ |
| AIMK | $O(n^2)$ |
| AIMK-RS | $O(n)$ |

## 4. Conclusion

In this paper, we proposed an adaptive initialization method for the K-means algorithm, which not only adapts to datasets with various characteristics but also requires only two runs to obtain better clustering results. In addition, we proposed the AIMK-RS based on random sampling to reduce the time complexity of the AIMK to O(n). AIMK-RS is easily applied to large and high-dimensional datasets. First, we proposed a new threshold to calculate the density of the data points based on the skeleton points of the MST. In addition, after using the new threshold, we found that we only need to adjust the parameter twice, and we can obtain better clustering results. Finally, we applied random sampling to AIMK to obtain the AIMK-RS, whose time complexity is only O(n).

In the future, we will use AIMK or AIMK-RS to initialize other variants of the K-means algorithm, such as the K-medoids algorithm, K-modes algorithm, fuzzy C-means algorithm, etc. In addition, the threshold *Thr* proposed in this paper can be used to help density-based clustering algorithms, such as DBSCAN (Ester et al., 1996), OPTICS (Ankerst et al., 1999), and SFDP, calculate the density of points without any extra adjusting parameters. Furthermore, we will continue to explore the new method to estimate the characteristics of datasets to further determine the parameter λ specifically.

## Acknowledgments

This work was supported in part by the Australian Research Council (ARC) under discovery grant DP180100670 and DP180100656. We also thank the NSW Defence Innovation Network and NSW State Government of Australia for financial support of this project through grant DINPP2019 S1-03/09. Research was also sponsored in part by the Office of Naval Research Global, US, and was accomplished under Cooperative Agreement Number ONRG - NICOP - N62909-19-1-2058.